\documentclass[11pt]{article}
\usepackage{latexsym}
\usepackage{graphicx}
\usepackage{epsfig}
\usepackage{url}
\usepackage{amsmath,amssymb}
\numberwithin{equation}{section}

\graphicspath{ {./figures/} }
\usepackage{hyperref}
\usepackage{float}
\usepackage{verbatim}
\usepackage{color}
\usepackage{amsmath}
\usepackage{amsfonts}

\usepackage{authblk}
\usepackage{algorithm}
\usepackage{algorithmic}
\usepackage{soul}
\usepackage{multirow}
\usepackage{color}
\usepackage{subcaption}
\usepackage[draft]{todonotes}   
\usepackage{enumerate}

\usepackage[utf8]{inputenc}
\usepackage[table]{xcolor}
\usepackage{hhline}
\usepackage{booktabs}
\usepackage{xspace}
\usepackage{caption} 
\usepackage{cleveref}

\newcommand{\name}{GradSkip\xspace} 

\begin{document}

\title{Gradient-Skipping Relevance Propagation for Efficient Explainability of Vision Transformers}

\author{Christopher Buratti$^1$, Michele Marchetti$^{1}$,Federica Parlapiano$^1$, Davide Traini$^{1, 2}$, Domenico Ursino$^1$, and Luca Virgili$^{1*}$ \\
$^{1}$ DII, Polytechnic University of Marche  \\
$^{2}$ CHIMOMO, University of Modena and Reggio Emilia\\
$^{*}$ Contact Author\\
c.buratti@pm.univpm.it; michele.marchetti@univpm.it; f.parlapiano@pm.univpm.it; davide.traini@unimore.it; d.ursino@univpm.it; luca.virgili@univpm.it}
\date{}

\maketitle

\begin{abstract}
Vision Transformers (ViTs) are difficult to interpret because current methods of relevance propagation and attention flow do not fully consider some key architectural features, such as the uneven importance of attention heads and residual connections. Prior approaches typically assume uniform importance across attention heads; furthermore, they model skip connections as identity paths, leading to inaccurate relevance attribution. To address these issues, we introduce \name{}, a novel relevance propagation method for ViTs based on adaptive head weighting and skip-aware propagation. \name{} models the different importance of the attention heads and dynamically distributes relevance between the attention and residual paths. Experiments on ImageNet1K and BloodMNIST demonstrate a state-of-the-art faithfulness of \name{} while requiring over 14 times fewer GFLOPs than the best-performing existing approaches. Additional evaluations using transformer-based segmentation confirm improved localization and alignment with ground-truth regions.
\end{abstract}


\section{Introduction}
\label{sec:Introduction}

Vision Transformers (ViTs) have emerged as a powerful alternative to Convolutional Neural Networks (CNNs) for many computer vision tasks, achieving state-of-the-art results on several benchmarks \cite{Waseem*26}. Unlike CNNs, which rely on local receptive fields, ViTs process images as sequences of patch tokens and use self-attention to capture long-range dependencies \cite{Vaswani*17, GoGoSi25}. While this design improves accuracy and scalability, it introduces challenges regarding interpretability. It is therefore crucial to identify which image regions influence a model’s predictions in order to make it possible to trust, debug, and understand learned representations. This need is particularly important in high-stakes fields, such as medical imaging, where understanding the rationale behind a prediction is essential for clinical trust and accountability, and where an opaque decision can have serious consequences.

ViT explainability methods can be categorized into three approaches: gradient-based, attention-based, and perturbation-based. Gradient-based approaches estimate feature importance using backpropagated gradients ~\cite{Selvaraju*17, Smilkov*17, AlSu22}. Attention-based approaches exploit self-attention weights to trace information flow across layers~\cite{Hao*21, Hafiz*21}. Perturbation-based approaches evaluate feature relevance by masking or modifying input regions and measuring the impact on model predictions \cite{Englebert*23,Xie*23,PeDaSa18}. While these approaches provide useful insights, they have limitations. These include unstable gradient attributions (for gradient-based approaches), limited causal reliability (for attention-based approaches), and  high computational costs (for perturbation-based approaches). Furthermore, most approaches treat attention heads uniformly and overlook residual connections, which hinders accurate relevance propagation in transformer architectures.

Fundamentally, these limitations stem from a shared issue in the propagation of relevance through the two mechanisms that define a transformer block, i.e., multi-head self-attention and residual connections \cite{Li*23-2}. Attention heads are known to play heterogeneous roles; some encode semantically focused patterns, while others attend diffusely across the input. Treating them uniformly (as most attention-based approaches do) mixes informative and uninformative signals, resulting in a diluted explanation. Residual connections allow a block to act as an identity mapping for certain tokens while significantly altering others. However, existing approaches usually handle the skip path implicitly or through fixed heuristics. This blurs the distinction between relevance attributed to the learned transformation and relevance simply preserved along the identity route. Consequently, the relevance signal is distributed inaccurately, and the explanation no longer reflects how the model processes its input.

To address these issues, we introduce \name{}, which employs two conceptual innovations. First, it develops an adaptive attention head weighting mechanism that combines gradient-based flow metrics and Gini-based sparsity measures to identify the most semantically relevant attention patterns. Second, it introduces skip-connection-aware relevance propagation, which  explicitly models the attention transformation path and residual connections during backward relevance computation. As experimental results show, on classification benchmarks \name{} achieves state-of-the-art faithfulness (measured using the Insertion - Deletion metric \cite{PeDaSa18} and the Faithfulness Violation Test \cite{Liu*22-2}) while maintaining highly competitive results on the Pointing Game localization metric \cite{Englebert*23}. Furthermore, for semantic segmentation on ADE20K-150, \name{} attains the highest Pixel Accuracy \cite{LoShDa15} and mean Intersection over Union (mIoU) \cite{LoShDa15}, outperforming the second-best approach by 53.94\% in mIoU. With only 69.39 Giga Floating Point Operations (GFLOPs) for classification and 28.52 GFLOPs for segmentation, \name{} matches the efficiency of LRP \cite{Bach*15} while surpassing perturbation baselines, enabling practical deployment. This efficiency is a key feature of \name{}, which delivers the faithfulness typically associated with costly perturbation-based approaches at a fraction of their computational budget. Finally, qualitative results demonstrate semantically coherent explanations across tasks.

The main contributions of this paper can be summarized as follows:

\begin{itemize}

    \item We propose an adaptive attention head weighting mechanism that combines a gradient-based flow metric and a Gini-based sparsity measure. This mechanism captures the causal influence of each head and the selectivity of its attention distribution. Thus, it moves beyond the uniform treatment of heads adopted by most existing approaches.
    
    \item We introduce a skip-connection-aware relevance propagation rule that dynamically partitions relevance between the attention transformation path and the residual path according to their relative causal contribution. This rule  explicitly respects the residual structure of transformer blocks, instead of relying on fixed heuristics.
    
    \item We conduct an extensive evaluation on classification benchmarks (ImageNet1K and BloodMNIST) and segmentation benchmarks (ADE20K-150), showing that \name{} achieves state-of-the-art faithfulness and localization scores, while requiring substantially fewer GFLOPs than competing methods.
    
\end{itemize}

The rest of this paper is organized as follows: Section~\ref{sec:Related} reviews related work.  Section~\ref{sec:Technical-Description-GradSkip} provides a technical description of \name{}. Section~\ref{sec:Experiments} reports the results of our experimental campaign. Section~\ref{sec:Discussion-section} provides a discussion on the strengths and limitations of \name{}. Finally, Section~\ref{sec:Conclusion} draws conclusions and outlines some possible directions for future work.

\section{Related Work}
\label{sec:Related}

ViT explainability approaches can be classified into three classes: gradient-based, attention-based, and perturbation-based.

Gradient-based approaches aim to identify regions of the input that most influence the model's predictions~\cite{Selvaraju*17, Smilkov*17, AlSu22}. They are based on the idea that the gradient of the output with respect to the input or intermediate features quantifies how sensitively the prediction responds to local changes and identifies the most important regions. Originally designed for CNNs, techniques like Grad-CAM~\cite{Selvaraju*17}, Grad-CAM++~\cite{Chattopadhay*18}, Score-CAM~\cite{Wang*20-3}, Integrated Gradients~\cite{SuTaYa17}, and SmoothGrad~\cite{Smilkov*17} have been extended to the ViT domain. Grad-CAM and its variants combine gradients with feature activations to produce class-discriminative maps. Integrated Gradients accumulates gradients along a path from a baseline to the input to satisfy attribution axioms. SmoothGrad averages gradients over noisy copies of the input to reduce visual noise. More complex attribution mechanisms, such as Shapley values~\cite{LuLe17} and Layer-wise Relevance Propagation (LRP)~\cite{Bach*15, HaCh25}, have required custom adaptations for transformer architectures. These adaptations include attention masking schemes and specialized propagation rules~\cite{CoKiLe23, Voita*19, ChGuWo21}. Following the initial LRP adaptations for ViTs~\cite{ChGuWo21}, several variants have been proposed to refine how relevance flows through self-attention layers. These include: {\em (i)} conservative propagation strategies that enforce stricter relevance conservation across deep network layers~\cite{Ali*22}, {\em (ii)} attention-aware formulations, like AttnLRP, which integrate the attention matrix into the backward propagation process~\cite{Achtibat*24}, and {\em (iii)} approaches that revisit LRP by incorporating positional attribution to better localize token significance~\cite{Bakish*25}. Recent work has introduced additional refinements, such as Finer-CAM~\cite{Zhang*25-1}, which augments Grad-CAM by highlighting discriminative regions in fine-grained tasks. The main strength of gradient-based approaches is their low computational cost since attributions are typically obtained with a single backward pass. Instead, their main weakness lies in the instability of raw gradients in deep transformers.

Attention-based approaches leverage the self-attention mechanism to trace information flow within the model~\cite{Hao*21, Hafiz*21}. Since self-attention explicitly encodes how each token aggregates information from the others, these approaches treat the attention weights as a natural and readily available explanation signal, eliminating the need for gradients or additional forward passes. Notable contributions in this area include the attention-flow and attention-rollout approaches~\cite{AbZu20}, where the propagation of input tokens is modeled using directed acyclic graphs and maximum flow algorithms to account for how attention composes across successive layers. Other frameworks, such as Transition Attention Maps (TAM)~\cite{Yuan*21}, conceptualize attention as a Markov process, modeling the layer-wise propagation of relevance as a sequence of state transitions. While attention-based approaches are intuitive and inexpensive, they aggregate the contribution of the different heads with uniform rules and rely on raw attention magnitudes. This limits the faithfulness of the resulting explanations.

Mask-based and perturbation-oriented approaches offer an alternative perspective. They systematically modify input features or token sets and measure the resulting impact on the model's decisions. Rather than examining the model's internal workings, they treat the network as a black box and infer the importance of a region by observing how much the prediction degrades when that region is removed or altered. Randomized Input Sampling for Explanation (RISE)~\cite{PeDaSa18} accomplishes this by applying random masks to input images and observing changes in predictions. It estimates saliency as the expected output over many masked variants. This concept has been adapted and extended for ViTs in recent works, including ViT-CX~\cite{Xie*23}, which aggregates final-layer embeddings to create saliency maps, and TIS~\cite{Englebert*23}, which strategically applies perturbations after linear projection and positional encoding. Another example is MUTEX~\cite{Ursino-NCAA-25}, which views attention matrices across layers as multilayer networks to generate masks that produce explanation maps. Perturbation-based approaches often yield highly faithful explanations because they directly probe model behavior; however, this comes at the cost of a large number of forward passes. This makes perturbation-based approaches substantially more expensive than the other two families of ViT explainability approaches.

Despite these advances, existing ViT explainability approaches have limitations. Gradient-based approaches may suffer from saturation and vanishing gradients in deep architectures. Attention-based approaches can yield misleading explanations because attention weights do not necessarily reflect causal relevance \cite{Bibal*22, JaWa19}. Finally, perturbation-based approaches are computationally expensive due to the large number of forward passes required. These limitations highlight the need for an approach that retains the efficiency of propagation-based attribution while explicitly accounting for the non-uniform role of attention heads and the residual structure of transformer blocks. \name{} aims to fill this gap.

\section{Technical Description of \name{}}
\label{sec:Technical-Description-GradSkip}

In this section, we provide a technical description of \name{}. Specifically, Subsection \ref{sub:Motivation} explains the rationale behind its architecture, while Subsection \ref{sub:GradSkip-Approach} describes the various steps that comprise its approach.

\subsection{Motivation}
\label{sub:Motivation}

A core obstacle to making ViT explainable is that the attribution signal must propagate through two essential, yet highly nonuniform, architectural mechanisms, i.e., multi-head self-attention and skip connections. Many existing approaches either aggregate heads with uniform rules (e.g., simple averaging or rollout)~\cite{AbZu20, Yuan*21}, implicitly ignoring the well-documented functional heterogeneity of different attention heads~\cite{Voita*19}, or rely on raw attention weights as a direct proxy for explanation. However, the literature has extensively demonstrated that high attention weights do not necessarily reflect true causal influence on the model's output~\cite{JaWa19, Bibal*22}. Additionally, residual paths are often handled implicitly~\cite{Bach*15} or via static heuristics \cite{ChGuWo21, Achtibat*24}, obscuring whether relevance should be assigned to the learned transformation or simply preserved along the identity route.

\name{}'s approach addresses the above-mentioned issues thanks to two main features:

\begin{itemize}

\item{\em Dual-perspective head weighting.} Attention heads play different roles and encode distinct structural and semantic patterns across layers~\cite{Voita*19}. Therefore, assuming that all heads are equally informative is a strong modeling assumption that injects noise into relevance propagation. \name{} addresses this issue by weighting the heads using two complementary perspectives: (\emph{i}) a gradient-driven notion of predictive influence, which quantifies the factors impacting the decision, and (\emph{ii}) a sparsity-driven notion of spatial–semantic focus, which identifies where attention concentrates. Combining these signals yields a dynamic head-selection mechanism that is intrinsically more robust than relying on either criterion alone. Gradients capture a head's causal importance for prediction, whereas sparsity reflects the selective distribution of head attention.

\item{\em Dynamic skip-aware relevance partitioning.} Transformer blocks compute outputs via residual composition. This means that a layer can act primarily as an identity mapping for certain input tokens while applying strong nonlinear transformations to others. Interpreting these blocks requires making a principled decision about how much relevance should flow through the main attention path versus skip connection. \name{} models skip connections using an input-dependent metric that partitions relevance according to the relative causal contribution of the two paths. It uses gradient-weighted activation magnitudes as a proxy for path impact. This yields a dynamic, skip-aware propagation rule that strictly respects the fundamental residual structure of the architecture. This procedure differs from existing transformer attribution methods, which rely on fixed heuristics for the residual path rather than data-driven gradient flow~\cite{ChGuWo21}.

\end{itemize}


\subsection{The approach underlying GradSkip}
\label{sub:GradSkip-Approach}

Let $T$ be a ViT model, with $L$ layers and a patch size $p \times p$, which is used for image classification. Let $I$ be an image with height $h_I$ and width $w_I$. Assume that $I$ belongs to class $c$ and is provided in input to $T$. $I$ can be divided into $n$ tokens, where $n = \frac{h_I}{p} \cdot \frac{w_I}{p}$ (we consider only the patch tokens and discard other possible tokens, such as the CLS and the distillation tokens). The goal of \name{} is to generate an explanation map $\mathbf{S}$ that represents the importance of the tokens of $I$ for $c$. In order to reach this goal, \name{} performs the steps described in the following subsections.

\subsubsection{Forward Pass with Gradient Capture}
\label{subsub:Gradient-Capture}

For the class $c$, \name{} performs a single forward pass through $T$, capturing the intermediate representations $\mathbf{O}^l \in \mathbb{R}^{n \times d}$, gradients $\nabla \mathbf{O}^l \in \mathbb{R}^{n \times d}$, and attention maps $\mathbf{A}^l \in \mathbb{R}^{h \times n \times n}$ at each layer $l \in \{1, \ldots, L\}$. Here, $d$ represents the embedding dimension, and $h$ is the number of attention heads.

\subsubsection{Gradient Rescaling}
\label{subsub:Gradient-Rescaling}

To address the gradient flow imbalances inherent in transformer architectures, \name{} applies normalization. For each layer $l$, it computes the normalized gradient $\tilde{\nabla}\,\mathbf{O}^l$ as follows:

\noindent
\begin{minipage}[t]{0.48\linewidth}
\vspace{0pt}
\[
\lambda^l = \frac{\mu}{\|\nabla \mathbf{O}^l\|_F + \epsilon},
\]
\end{minipage}\hspace{6mm}
\begin{minipage}[t]{0.48\linewidth}
\vspace{0pt}
\begin{equation}
\tilde{\nabla}\,\mathbf{O}^l = \lambda^l \cdot \nabla \mathbf{O}^l
\label{eq:rescaled_gradients}
\end{equation}
\end{minipage}
\par\vspace{\belowdisplayskip}

Here, $\lambda^l$ is the balancing coefficient, $\|\nabla \mathbf{O}^l\|_F$ is the Frobenius norm \cite{Mirsky60} of the gradients at the layer $l$, $\mu = \frac{1}{L} \sum_{l=1}^{L} \|\nabla \mathbf{O}^l\|_F$ is the mean Frobenius norm of the gradients across all layers, and $\epsilon = 10^{-12}$ is a numerical stability constant. 

This normalization promotes FullGrad-completeness \cite{MeBaPi25}, a property that CNNs naturally satisfy but transformers violate due to non-locally-affine operations~\cite{SrFl19}.

\subsubsection{Initial Relevance Computation}
\label{subsub:Initial-Relevance-Computation}

\name{} calculates the relevance distribution $\mathbf{R}^L_i$ in the last layer \cite{Otsuki*24}, and backpropagates it to the original input features. In particular, it computes $\mathbf{R}^L_i$ using gradient-weighted activations as follows:

\begin{equation}
\mathbf{R}^L_i = \frac{\sum_{k=1}^{d} |\tilde{\nabla} O^L_{i_k}| \cdot |O^L_{i_k}|}{\sum_{j=1}^{n} \sum_{k=1}^{d} |\tilde{\nabla} O^L_{j_k}| \cdot |O^L_{j_k}|}
\end{equation}

Here, $i \in \{1, \ldots, n\}$ indexes the patch tokens. This initialization strategy weights each token's contribution based on both gradient magnitude and activation strength \cite{Barkan*21}.

\subsubsection{Adaptive Attention Head Weighting}
\label{subsub:Adaptive-Attention-Head-Weighting}

Unlike other methods that treat all attention heads uniformly, \name{} computes head-specific importance weights. To this end, it employs two complementary metrics that capture both gradient-based relevance and attention distribution characteristics. First, we examine the two metrics, and then we show how \name{} combines them.

\paragraph {Gradient-Based Flow Metric}
\ 

For each head $q \in \{1, \ldots, h\}$ at the layer $l$, \name{} computes:

\begin{equation}
g_q^l = \sum_{i=1}^n \sum_{j=1}^n \mathbf{A}^l_{q_{i_j}} \cdot \|\tilde{\nabla} \mathbf{O}^l_j\|_2
\end{equation}

This metric quantifies the gradient magnitude weighted by attention scores for each head. It effectively measures the aggregated gradient-weighted attention score of the head. To filter out noise and retain only the heads with substantial gradient flow, \name{} applies a threshold obtaining $\hat{g}_q^l$ defined as:

\begin{equation}
\hat{g}_q^l = \begin{cases}
g_q^l & \text{if } g_q^l \geq \gamma \cdot \max (g_{q'}^l) \\
0 & \text{otherwise}
\end{cases}
\label{eq:grad-flow}
\end{equation}

Here, $\gamma \in [0, 1]$ is a threshold hyperparameter, and $\max(g_{q'}^l)$ is the maximum value among all the heads in the layer $l$. 

Then, \name{} linearly normalizes the gradient-based weights $\hat{g}_q^l$, obtaining the normalized parameters $\tilde{g}_{q}^{l}$, which satisfy $\sum_{q=1}^{h} \tilde{g}_{q}^{l} = 1$.

\paragraph{Gini-Based Sparsity Metric}
\

To capture attention distribution patterns, \name{} computes the Gini coefficient \cite{Farris10} for each head. Let $a^q = \{a_1^q, \cdots, a_{n^2}^q\}$ denote the attention values of the head $q$ sorted in ascending order. \name{} calculates the Gini coefficients $G_q$ and the corresponding normalized sparsity-based weights $\tilde{s}_q^l$ as follows:

\ 

\noindent
\begin{minipage}[c]{0.56\linewidth}
  \centering
  $\displaystyle G_q = \frac{2 \sum_{u=1}^{n^2} u \cdot a_u^q}{n^2 \sum_{u=1}^{n^2} a_u^q} - \frac{n^2 + 1}{n^2} ,$
\end{minipage}\hfill
\begin{minipage}[c]{0.40\linewidth}
  \begin{equation}
    \tilde{s}_q^l = \frac{G_q}{\sum_{q=1}^{h} G_{q} + \epsilon}
    \label{eq:gini}
  \end{equation}
\end{minipage}
\par\vspace{\belowdisplayskip}

The numerical stability constant $\epsilon$ is the same one introduced in Equation \ref{eq:rescaled_gradients}. High Gini coefficients indicate spiked attention patterns, which often correspond to focused, semantically meaningful regions, distinguishing them from heads with diffuse attention distributions. We point out that we used the Gini coefficient instead of other metrics (e.g., Shannon entropy \cite{Shannon48}) because it inherently produces values normalized in the range $[0, 1]$, enabling direct fusion with gradient-based weights.

\paragraph{Flow-Sparsity Fusion}
\

The final weight for a head $q$ is determined by combining both metrics through a convex combination:

\begin{equation}
\omega_q^l = \alpha \cdot \tilde{g}_q^l + (1 - \alpha) \cdot \tilde{s}_q^l
\label{eq:head-weight}
\end{equation}

Here, $\alpha \in [0, 1]$ is a hyperparameter coefficient. This linear interpolation balances causal relevance, captured by gradient flow, and spatial coherence, captured by sparsity. As demonstrated in our ablation study (\Cref{tab:Hyperparameter-alpha}), relying solely on one signal ($\alpha=0$ or $\alpha=1$) results in suboptimal performance. This confirms that these metrics provide complementary information that, when combined, maximizes explanation faithfulness. 
The weights $\omega_q^l$ are renormalized, yielding $\tilde{\omega}_{q}^{l}$ such that $\sum_{q=1}^{h} \tilde{\omega}_{q}^{l} = 1$.

\subsubsection{Skip-Connection-Aware Relevance Propagation}
\label{subsub:Relevance-Propagation}

\name{} propagates relevance backward through the ViT by explicitly modeling both the attention transformation path and skip connections present in transformer blocks.


At the layer $l$, it constructs a weighted attention matrix $\mathbf{W}^l$ by combining head-weighted attention with gradient and embedding magnitudes as follows:

\begin{equation}
\mathbf{W}^l_{i_j} = \left(\sum_{q=1}^{h} \tilde{\omega}_q^l \mathbf{A}^l_{q_{i_j}}\right) \cdot \|\tilde{\nabla} \mathbf{O}^l_i\|_2 \cdot \|\mathbf{O}^l_j\|_2
\end{equation}

Here, $\tilde{\omega}_q^l$ are the head importance weights defined in Section \ref{subsub:Adaptive-Attention-Head-Weighting}, $\|\tilde{\nabla} \mathbf{O}^l_i\|_2$ captures the rescaled gradient magnitude at the querying token $i$, and $\|\mathbf{O}^l_j\|_2$ captures the embedding magnitude at the attended token $j$. 
The matrix is then row-normalized, yielding $\tilde{\mathbf{W}}^{l}$ such that $\sum_{j=1}^{n} \tilde{\mathbf{W}}^{l}_{i_j} = 1$.


To account for residual connections in transformer blocks, \name{} computes the relative contribution of the main attention and the skip connection paths by comparing their aggregate causal impacts. The causal impact of a path is quantified as its total absolute gradient-weighted activation and is computed as:

\

\noindent
\begin{minipage}[c]{0.68\linewidth}
  \centering
  $\displaystyle \beta_{main}^l =
  \frac{
  \sum_{i=1}^n \sum_{k=1}^d \left| \tilde{\nabla} O^l_{i_k} \cdot O^l_{i_k} \right|
  }{
  \sum_{i=1}^n \sum_{k=1}^d \left(
  \left| \tilde{\nabla} O^l_{i_k} \cdot O^l_{i_k} \right|
  + 
  \left| \tilde{\nabla} O^l_{i_k} \cdot O^{l-1}_{i_k} \right|
  \right)
  } , $
\end{minipage}\hfill
\begin{minipage}[c]{0.30\linewidth}
  \begin{equation}
  \beta_{skip}^l = 1 - \beta_{main}^l
  \label{eq:beta_skip}
  \end{equation}
\end{minipage}
\par\vspace{\belowdisplayskip}

Here, the numerator of $\beta_{main}^l$ quantifies the causal relevance of the main transformation path (the output $O^l$ of the block), while the denominator represents the total causal relevance generated by the residual block as a whole, including the main and skip paths (represented by $O^{l-1}$). Thus, the coefficient $\beta_{main}^l$ acts as a normalized relevance gate that dynamically modulates the backward flow of relevance. Instead, the coefficient of $\beta_{skip}^l$ acts as a complementary normalized gate that dynamically preserves relevance through the skip connection. This formulation ensures that blocks predominantly acting as identity mappings contribute minimally to the explanation, with most relevance flowing through the skip connection.

The relevance at the layer $l-1$ is computed by combining contributions from both paths:

\begin{equation}
\mathbf{R}^{l-1} = \beta_{main}^l \cdot (\tilde{\mathbf{W}}^l)^T \mathbf{R}^l + \beta_{skip}^l \cdot \mathbf{R}^l
\end{equation}

Here, the first term represents the relevance propagated through the attention mechanism, while the second term represents the relevance propagated through the skip connection. 

To ensure the conservation of approximate relevance across layers, which is a desirable property for faithful attribution \cite{Otsuki*24}, we apply normalization as follows:

\begin{equation}
\tilde{\mathbf{R}}^{l-1} = \mathbf{R}^{l-1} \cdot \frac{\sum_{i=1}^{n} \mathbf{R}^l_i}{\sum_{i=1}^{n} \mathbf{R}^{l-1}_i + \epsilon}
\end{equation}

Again, the numerical stability constant $\epsilon$ is the same one introduced in Equation \ref{eq:rescaled_gradients}. 

This process is repeated by decreasing $l$ from $L$ down to 2, yielding a sequence of relevance distributions $\{\tilde{\mathbf{R}}^L, \tilde{\mathbf{R}}^{L-1}, \ldots, \tilde{\mathbf{R}}^1\}$, each belonging to $\mathbb{R}^n$. The final saliency map is derived from the relevance $\tilde{\mathbf{R}}^1$ of the first layer. This relevance is reshaped into a two-dimensional spatial grid to obtain the explanation map $\mathbf{S}$ denoting the importance of each token of the image $I$ for the class $c$.

\section{Experiments}
\label{sec:Experiments}

This section describes the experimental campaign that we conducted to validate \name{}. Specifically, Subsection \ref{sub:Experimental-Setup} presents the experimental setup. Subsection \ref{sub:Quantitative-Results} illustrates the quantitative results. Finally, Subsection \ref{sub:Qualitative} describes the qualitative results.

\subsection{Experimental Setup}
\label{sub:Experimental-Setup}

In our experiments, we used the Vision Transformer (ViT) \cite{Dosovitskiy*21} and the Data-Efficient Image Transformer (DeiT) \cite{Touvron*21} systems as the reference ViT models for classification, and the SegFormer system \cite{Xie*21-3} as the reference ViT model for segmentation.

For comparison with \name{}, we used MUTEX~\cite{Ursino-NCAA-25}, TIS~\cite{Englebert*23}, ViT-CX~\cite{Xie*23}, Chefer1~\cite{ChGuWo21}, RISE \cite{PeDaSa18}, Finer-CAM~\cite{Zhang*25-1}, {\color{black} Pe-LRP~\cite{Bakish*25}, AH+LN~\cite{Ali*22} and AttnLRP~\cite{Achtibat*24}}. We evaluated all baseline approaches using the optimal hyperparameter values obtained through a hyperparameter tuning activity. For \name{}, we set $\gamma$ to $0.25$ and $\alpha$ to $0.50$; we selected these values via systematic hyperparameter optimization (see Section~\ref{subsub:hyperparameters_analysis}).

For classification, we tested \name{} on two different datasets, following the assessment protocols used in previous works on visual explainability for this task \cite{Xie*23, Chen*22-2}. The first dataset is a random subset of the ImageNet1K validation set \cite{Russakovsky*15} consisting of 5,000 images \cite{Xie*23, Chen*22-2}. The second dataset is BloodMNIST \cite{Yang*23}, which is from the medical domain. For segmentation, we used the widely adopted ADE20K-150 dataset \cite{Zhou*17-1}. 

For classification, we evaluated faithfulness using the Insertion–Deletion metric \cite{PeDaSa18} and the Faithfulness Violation Test \cite{Liu*22-2}. Insertion - Deletion is computed by progressively inserting or removing pixels in descending order of relevance and tracking the resulting change in model output. The Faithfulness Violation Test detects inconsistencies in saliency maps by quantifying how often the relevance assigned to a feature contradicts its true contribution to the model's output. We employed multiple perturbation strategies (i.e., Mean, Black, Random) for replacing pixels to ensure robustness across different obfuscation schemes. Additionally, we measured spatial accuracy using the Pointing Game localization metric \cite{Englebert*23}. For the segmentation task, following other approaches \cite{Bakish*25}, we binarized the generated heatmap and selected only the 75\% most relevant pixels. We then used this heatmap to calculate two complementary metrics: Pixel Accuracy (PA) \cite{LoShDa15} and mean Intersection over Union (mIoU) \cite{LoShDa15}. PA measures the fraction of correctly classified pixels. mIoU computes the overlap between the prediction and the ground truth, normalized by their union. Finally, we performed an efficiency analysis by measuring the Giga Floating Point Operations (GFLOPs) required for each approach, as well as a qualitative analysis on different images.

\subsection{Quantitative Results}
\label{sub:Quantitative-Results}

In this section, we present the quantitative results from our experimental campaign. Specifically, Subsection \ref{subsub:Classification-Metrics} illustrates the quantitative results concerning classification. Subsection \ref{subsub:Segmentation-Metrics} analyzes the quantitative results regarding segmentation. Subsection \ref{subsub:ablation_analysis} presents an ablation study related to \name{}. Finally, Subsection \ref{subsub:hyperparameters_analysis} proposes a hyperparameter analysis related to our framework.

\subsubsection{Classification}
\label{subsub:Classification-Metrics}

\paragraph{Faithfulness Metrics}
\

First, we computed the Insertion - Deletion metric and the Faithfulness Violation Test values on Vit-Base and DeiT-Base using Mean, Black, and Random perturbation strategies across ImageNet1K and BloodMNIST. Higher Insertion - Deletion values and lower Faithfulness Violation Test values reflect a more faithful explanation. The obtained values are reported in Table \ref{tab:Insertion - Deletion-ViolationTest}. In this and subsequent tables, an up (resp., down) arrow near the metric name indicates that higher (resp., lower) values are better for the framework's performance. In each column, the optimal value is in bold and the suboptimal value is underlined.

As shown in Table \ref{tab:Insertion - Deletion-ViolationTest}, \name{} achieves strong performance in all evaluation metrics when applied to the ViT-Base model. On ImageNet1K, for example, \name{} outperforms the second-best approach (MUTEX) by 2.33\%, 4.65\%, and 12.50\% using the Black, Mean, and Random perturbation strategies respectively, when considering the Insertion - Deletion metric. TIS ranks third. Regarding the Faithfulness Violation Test, \name{} improves by 21.43\% upon MUTEX and 31.25\% upon TIS using the Mean perturbation, indicating markedly enhanced robustness. It improves by 20.00\% upon MUTEX using the Black perturbation, though it falls slightly behind Finer-CAM by 7.69\% using the Random perturbation. This performance consistency extends to BloodMNIST, where \name{} matches or surpasses MUTEX and the other baselines in both the Insertion - Deletion metric and the Faithfulness Violation Test.

\begin{table}[ht!]
\footnotesize
\centering
\resizebox{0.95\textwidth}{!}{
\begin{tabular}{l|l|ccc|ccc|ccc|ccc}
\toprule
 & & \multicolumn{6}{c|}{{\em ImageNet1K}} & \multicolumn{6}{c}{{\em BloodMNIST}} \\
\cmidrule(l){3-14}
&  & \multicolumn{3}{c|}{{\em Insertion - Deletion} ↑} & \multicolumn{3}{c|}{{\em Violation Test} ↓} & \multicolumn{3}{c|}{{\em Insertion - Deletion} ↑} & \multicolumn{3}{c}{{\em Violation Test} ↓} \\
\cmidrule(l){3-14}
{\em Model}  & {\em Method} & {\em Mean} & {\em Black} & {\em Rand} & {\em Mean} & {\em Black} & {\em Rand} & {\em Mean} & {\em Black} & {\em Rand} & {\em Mean} & {\em Black} & {\em Rand} \\
\midrule
\multirow{11}{*}{ViT-Base} 
 & \name{}  & \textbf{0.45} & \textbf{0.44} & \textbf{0.45} & \textbf{0.11} & \textbf{0.12} & \underline{0.14} & \textbf{0.49} & \textbf{0.47} & \textbf{0.39} & \textbf{0.12} & \textbf{0.11} & \textbf{0.13} \\
 & MUTEX \cite{Ursino-NCAA-25}& \underline{0.43} & \underline{0.43} & \underline{0.40} & \underline{0.14} & \underline{0.15} & \underline{0.14} & \underline{0.46} & \underline{0.46} & \underline{0.38} & 0.18 & \underline{0.13} & \underline{0.18} \\
 & TIS~\cite{Englebert*23} & 0.42 & 0.40 & 0.38 & 0.16 & 0.18 & 0.15 & 0.44 & 0.45 & \underline{0.38} & \underline{0.15} & \underline{0.13} & 0.19 \\
 & ViT-CX~\cite{Xie*23} & 0.28 & 0.31 & 0.35 & 0.19 & 0.22 & 0.17 & 0.18 & 0.16 & 0.20 & 0.22 & 0.21 & \textbf{0.13} \\
 & Chefer1~\cite{ChGuWo21} & 0.29 & 0.29 & 0.29 & 0.21 & 0.23 & 0.19 & 0.21 & 0.25 & 0.23 & 0.24 & 0.22 & 0.19 \\
 & TAM~\cite{Yuan*21} & 0.27 & 0.26 & 0.26 & 0.22 & 0.24 & 0.20 & 0.16 & 0.20 & 0.23 & 0.25 & 0.23 & 0.21 \\
 & RISE~\cite{PeDaSa18} & 0.30 & 0.30 & 0.31 & 0.20 & 0.22 & 0.18 & 0.22 & 0.29 & 0.24 & 0.23 & 0.20 & 0.19 \\
 & Finer-CAM~\cite{Zhang*25-1} & 0.37 & 0.37 & 0.31 & 0.18 & 0.19 & \textbf{0.13} & 0.23 & 0.23 & 0.25 & 0.20 & 0.19 & \textbf{0.13} \\
 & Pe-LRP~\cite{Bakish*25}    & 0.32 & 0.29 & 0.33 & 0.19 & 0.23 & 0.17 & 0.30 & 0.28 & 0.27 & 0.22 & 0.20 & 0.21 \\
 & AH+LN~\cite{Ali*22}        & 0.26 & 0.28 & 0.25 & 0.25 & 0.23 & 0.24 & 0.24 & 0.27 & 0.20 & 0.28 & 0.26 & 0.25 \\
 & AttnLRP~\cite{Achtibat*24} & 0.29 & 0.26 & 0.30 & 0.23 & 0.27 & 0.20 & 0.27 & 0.24 & 0.23 & 0.26 & 0.27 & 0.24 \\
\midrule
\multirow{11}{*}{DeiT-Base} 
 & \name{}  & \textbf{0.47} & \textbf{0.49} & \textbf{0.46} & \textbf{0.12} & \textbf{0.13} & \textbf{0.12} & \textbf{0.49} & \textbf{0.49} & \textbf{0.48} & \textbf{0.14} & \textbf{0.13} & \textbf{0.16} \\
 & MUTEX \cite{Ursino-NCAA-25}& \underline{0.44} & \underline{0.44} & \underline{0.44} & \underline{0.15} & \underline{0.16} & 0.15 & \underline{0.48} & \underline{0.48} & \textbf{0.48} & 0.18 & 0.19 & 0.19 \\
 & TIS~\cite{Englebert*23} & 0.42 & 0.42 & 0.41 & 0.17 & 0.19 & 0.16 & 0.47 & 0.47 & \underline{0.46} & \underline{0.15} & 0.16 & \underline{0.17} \\
 & ViT-CX~\cite{Xie*23} & 0.31 & 0.31 & 0.30 & 0.20 & 0.23 & 0.18 & 0.28 & 0.24 & 0.22 & 0.23 & 0.21 & 0.20 \\
 & Chefer1~\cite{ChGuWo21} & 0.29 & 0.29 & 0.29 & 0.22 & 0.24 & 0.20 & 0.27 & 0.27 & 0.23 & 0.25 & 0.23 & 0.22 \\
 & TAM~\cite{Yuan*21} & 0.27 & 0.26 & 0.26 & 0.23 & 0.25 & 0.21 & 0.28 & 0.24 & 0.27 & 0.26 & 0.24 & 0.23 \\
 & RISE~\cite{PeDaSa18} & 0.30 & 0.30 & 0.31 & 0.21 & 0.23 & 0.19 & 0.27 & 0.30 & 0.22 & 0.24 & 0.22 & 0.20 \\
 & Finer-CAM~\cite{Zhang*25-1} & 0.37 & 0.37 & 0.31 & 0.19 & 0.20 & \underline{0.14} & 0.28 & 0.28 & 0.27 & 0.23 & \underline{0.15} & 0.18 \\
 & Pe-LRP~\cite{Bakish*25}    & 0.35 & 0.31 & 0.34 & 0.17 & 0.21 & 0.16 & 0.33 & 0.29 & 0.28 & 0.20 & 0.18 & 0.20 \\
 & AH+LN~\cite{Ali*22}        & 0.28 & 0.30 & 0.27 & 0.23 & 0.20 & 0.22 & 0.26 & 0.29 & 0.22 & 0.25 & 0.24 & 0.23 \\
 & AttnLRP~\cite{Achtibat*24} & 0.31 & 0.29 & 0.32 & 0.21 & 0.24 & 0.18 & 0.30 & 0.26 & 0.25 & 0.24 & 0.25 & 0.22 \\
\bottomrule

\end{tabular}
}
\caption{Results of the computation of the Insertion - Deletion metric and Faithfulness Violation Test on ViT-Base and DeiT-Base across ImageNet1K and BloodMNIST using Mean, Black and Random perturbation strategies}
\label{tab:Insertion - Deletion-ViolationTest}
\end{table}

Table \ref{tab:Insertion - Deletion-ViolationTest} also shows that \name{} maintains superior performance on DeiT-Base across ImageNet1K. For the Insertion - Deletion metric, \name{} outperforms MUTEX by 6.82\% and TIS by 11.90\% using the Mean perturbation. Using the Black perturbation, it improves by 11.36\% upon MUTEX. Using the Random perturbation, \name{} outperforms MUTEX by 4.55\% while exceeding TIS's performance by 12.20\%. For the Faithfulness Violation Test on ImageNet1K, \name{} outperforms MUTEX by 20.00\% using the Mean perturbation and by 18.75\% using the Black perturbation. It achieves the best performance using the Random perturbation, improving by 20.00\% upon MUTEX. Even with the DeiT-Base model, the performance on the BloodMNIST dataset is consistent with that on ImageNet1K. 

These results highlight the effectiveness and stability of \name{} across different domains and models.

\paragraph{Localization Metric}
\ 

As outlined in Section \ref{sub:Experimental-Setup}, we used the Pointing Game metric to measure spatial accuracy. Specifically, we calculated the Pointing Game values on ViT-Base and DeiT-Base across the ImageNet1K dataset. We were unable to calculate this metric on BloodMNIST because it does not contain the bounding boxes necessary for this task. Higher Pointing Game values indicate better spatial accuracy. The obtained results are illustrated in Table \ref{tab:PG_comparison}.

\begin{table}[ht!]
\centering
    \resizebox{0.35\linewidth}{!}{%
    \begin{tabular}{lcc}
        \toprule
        \textit{Method}  & \textit{ViT-Base} ↑ & \textit{DeiT-Base} ↑ \\
        \midrule
        \name{}  & \textbf{0.868} & \underline{0.831} \\
        MUTEX~\cite{Ursino-NCAA-25} & \underline{0.861} & \textbf{0.837} \\
        TIS~\cite{Englebert*23} & 0.823 & 0.825 \\
        ViT-CX~\cite{Xie*23}  & 0.700 & 0.700 \\
        Chefer1~\cite{ChGuWo21} & 0.768 & 0.748 \\
        TAM~\cite{Yuan*21} & 0.737 & 0.635 \\
        RISE~\cite{PeDaSa18} & 0.753 & 0.766 \\
        Finer-CAM~\cite{Zhang*25-1} & 0.805  & 0.784 \\
        Pe-LRP~\cite{Bakish*25} & 0.614 & 0.633 \\
        AH+LN~\cite{Ali*22} & 0.582 & 0.521  \\
        AttnLRP~\cite{Achtibat*24} & 0.484 & 0.472 \\
    
        \bottomrule
    \end{tabular}
    }
    \caption{Results of the computation of the Pointing Game metric on ViT-Base and DeiT-Base across ImageNet1K}
    \label{tab:PG_comparison}
\end{table}

As shown in this table, \name{} achieves the best Pointing Game values on ViT-Base. It outperforms TIS by 5.47\% and maintains a slight edge over MUTEX. On DeiT-Base, \name{} trails MUTEX by just 0.72\% and continues to outperform TIS by 0.73\%. This consistency across architectures highlights the robustness of \name{}'s localization capability and suggests stable spatial reasoning across different models.

We point out that, although MUTEX and TIS achieve competitive localization results with respect to \name{}, they are severely limited by their high computational cost. Indeed, Table~\ref{tab:GFLOPS-Comparison} shows the GFLOPs required by each framework to generate a single attribution map using ViT-Base across the ImageNet1K dataset. This table reveals that \name{} requires only 69.39 GFLOPs. This is a remarkable reduction of 14.45 times in computational cost compared to MUTEX, which is the second best approach in faithfulness and the best in localization with the DeiT-Base model. \name{} demonstrates an even more substantial efficiency gain compared to TIS, requiring 256.66 times fewer operations. LRP-based approaches are the only ones that are slightly more efficient than \name{}. However, \name{} outperforms the best competing LRP-based approach (Pe-LRP) by 41.37\% on ViT-Base and by 31.28\% on DeiT-Base using the Pointing Game metric.

\begin{table}[ht!]
\centering
    \resizebox{0.25\linewidth}{!}{%
    \begin{tabular}{l|r}
        \toprule
        {\em Method} & {\em GFLOPs} ↓\\ 
        \midrule
        \name{} & 69.39 \\
        MUTEX~\cite{Ursino-NCAA-25} & 1,002.55 \\
        TIS~\cite{Englebert*23} & 17,808.70 \\
        ViT-CX~\cite{Xie*23} & 4,918.71 \\ 
        Chefer1~\cite{ChGuWo21} & 377.67 \\ 
        TAM~\cite{Yuan*21} & 2,203.77 \\ 
        RISE~\cite{PeDaSa18} & 140,927.36 \\ 
        Finer-CAM~\cite{Zhang*25-1} & 85.06 \\
        Pe-LRP~\cite{Bakish*25} & 68.21 \\
        AH+LN~\cite{Ali*22} & \underline{67.18}  \\
        AttnLRP~\cite{Achtibat*24} & \textbf{67.16} \\

        \bottomrule
    \end{tabular}
    }
    \caption{GFLOPs required to generate a single attribution map using ViT-Base across ImageNet1K}
    \label{tab:GFLOPS-Comparison}
\end{table}

These results demonstrate that \name{} strikes the optimal balance between the quality of its explanations and computational efficiency. This makes \name{} well suited for practical applications where both accuracy and computational resources are critical.

\subsubsection{Segmentation}
\label{subsub:Segmentation-Metrics}

Following the evaluation protocol described in Section \ref{sub:Experimental-Setup}, we assessed the segmentation performance of \name{} on the ADE20K-150 dataset using SegFormer as the reference model. We computed the PA and mIoU values and report them in Table ~\ref{tab:segmentation_results}. Here, higher values indicate better segmentation performance. 

\begin{table}[ht!]
\centering
    \resizebox{0.35\linewidth}{!}{%
    \begin{tabular}{lcc}
            \toprule
            \textit{Method} & \textit{PA} ↑ & \textit{mIoU} ↑ \\
            \midrule
            \name{} & \textbf{0.815} & \textbf{0.371} \\
            MUTEX~\cite{Ursino-NCAA-25} & 0.768 & \underline{0.241} \\
            TIS~\cite{Englebert*23} & 0.539 & 0.083 \\
            ViT-CX~\cite{Xie*23}  & 0.487 & 0.103 \\
            Chefer1~\cite{ChGuWo21} & 0.512 & 0.147 \\
            TAM~\cite{Yuan*21} & 0.641 & 0.228 \\
            RISE~\cite{PeDaSa18} & 0.527 & 0.121 \\
            Finer-CAM~\cite{Zhang*25-1} & \underline{0.803} & 0.104 \\
            Pe-LRP~\cite{Bakish*25} & 0.516 & 0.175 \\
            AH+LN~\cite{Ali*22} & 0.443 & 0.118  \\
            AttnLRP~\cite{Achtibat*24} & 0.386 & 0.163 \\
            \bottomrule
        \end{tabular}
    } 
    \caption{Results of the computation of PA and mIoU metrics on SegFormer across the ADE20K-150 dataset}
    \label{tab:segmentation_results}
\end{table}
    
As this table shows, \name{} achieves the best overall performance, with the highest PA and substantially higher mIoU than all competing approaches. Specifically, it outperforms MUTEX by 6.12\% in PA and by a significant 53.94\% in mIoU. Notably, while Finer-CAM achieves a comparable PA to \name{} (with a marginal difference of only 1.49\%), it yields a markedly lower mIoU; indeed, \name{} shows a 256.73\% improvement. This discrepancy occurs because Finer-CAM produces over-extended maps that assign high saliency values to large portions of the image. While this captures the target pixels (high PA), it also introduces significant noise by including irrelevant regions (low specificity). In contrast, \name{} maintains a high mIoU, demonstrating a superior ability to delineate class-specific regions precisely, while preventing over-assignment to the background and irrelevant areas.

Furthermore, as shown in Table~\ref{tab:GFLOPS-Comparison-segmentation}, \name{} is one of the most efficient approaches, requiring only 28.52 GFLOPs. Only other LRP-based methods (Pe-LRP, AH+LN, and AttnLRP) are more efficient, by a slight difference of no more than 0.61 GFLOPs. However, as shown in Table  \ref{tab:segmentation_results}, these approaches yield significantly poorer segmentation results than \name{}. In fact, compared to the most effective LRP alternative (e.g., Pe-LRP), \name{} achieves a substantial 57.95\% improvement in PA and a remarkable 112.00\% increase in mIoU.

\begin{table}[ht!]
\centering
    \resizebox{0.30\linewidth}{!}{%
    \begin{tabular}{l|r}
        \toprule
        {\em Method} & {\em GFLOPs} ↓ \\ 
        \midrule
        \name{} & 28.52 \\
        MUTEX~\cite{Ursino-NCAA-25} & 472.18 \\
        TIS~\cite{Englebert*23} & 8,423.81 \\
        ViT-CX~\cite{Xie*23} & 2,264.24 \\ 
        Chefer1~\cite{ChGuWo21} & 178.12 \\ 
        TAM~\cite{Yuan*21} & 1,015.03 \\ 
        RISE~\cite{PeDaSa18} & 65,746.37 \\ 
        Finer-CAM~\cite{Zhang*25-1} & 40.26 \\
        Pe-LRP~\cite{Bakish*25} & \underline{28.12} \\
        AH+LN~\cite{Ali*22} & \textbf{27.91}  \\
        AttnLRP~\cite{Achtibat*24} & \textbf{27.91} \\
        \bottomrule
    \end{tabular}
    }
    \caption{GFLOPs required to generate a single attribution map using SegFormer across ADE20K-150}
    \label{tab:GFLOPS-Comparison-segmentation}
\end{table}

\subsubsection{Ablation Study}
\label{subsub:ablation_analysis}

In this section, we empirically validate the role of attention sparsity in our head-selection strategy by analyzing its relationship with the Insertion - Deletion faithfulness metric. Specifically, we measure head sparsity using two indices: the Gini coefficient \cite{Farris10} and the Shannon Entropy \cite{Shannon48}, with the most focused heads having high Gini coefficients and low entropy values. Table~\ref{tab:statistical_tests} reports the results of the Spearman correlation and ANOVA tests between the attention sparsity indices and the Insertion - Deletion scores. They demonstrate a statistically significant correlation between the attention sparsity indices and the Insertion - Deletion metric. 

\begin{table}[ht!]
\centering
    \resizebox{0.30\linewidth}{!}{%
    \begin{tabular}{lcc}
        \toprule
        \textit{Test} & \textit{Gini} & \textit{Entropy} \\
        \midrule
        Spearman $r$ & 0.3870 & 0.3676 \\
        Spearman $p$ & $< 0.001$ & $< 0.001$ \\
        ANOVA $F$    & 388.14 & 340.09 \\
        ANOVA $p$    & $< 0.001$ & $< 0.001$ \\
        \bottomrule
    \end{tabular}
    }
    \caption{Correlation between attention sparsity metrics and Insertion - Deletion scores}
    \label{tab:statistical_tests}
\end{table}

Table~\ref{tab:means_by_percentile} shows the increase in the Insertion - Deletion performance gain obtained by using only the most relevant heads (i.e., those with high Gini coefficients and low entropy values) compared to the results obtained using all attention heads. The values in the table confirm that selecting heads from the highest sparsity percentiles (0\%-25\%) substantially improves performance compared to the all-heads baseline. In contrast, using heads from the lowest percentiles (75\%-100\%) results in a performance drop. Table~\ref{tab:means_by_percentile} also shows that the Gini coefficient outperforms entropy in providing a greater performance increment in the top quartile. Notably, both metrics demonstrate significant separation between the top and bottom quartiles, as indicated by Wilcoxon tests \cite{ReNe11}.

\begin{table}[ht!]
\centering
    \resizebox{0.40\linewidth}{!}{%
    \begin{tabular}{lcc}
        \toprule
        \textit{Percentile} & \textit{Gini} & \textit{Entropy} \\
        \midrule
        0\%-25\%    & $+0.264 \pm 0.210$ & $+0.252 \pm 0.250$ \\
        25\%-50\%   & $+0.057 \pm 0.203$ & $+0.044 \pm 0.204$ \\
        50\%-75\%   & $-0.025 \pm 0.179$ & $+0.026 \pm 0.186$ \\
        75\%-100\%  & $-0.091 \pm 0.173$ & $-0.038 \pm 0.184$ \\
        \midrule
        Wilcoxon & $< 0.001$ & $< 0.001$ \\
        \bottomrule
    \end{tabular}
    }
\caption{Average Insertion - Deletion increment obtained by using attention heads ranked by the Gini coefficient or the Shannon Entropy instead of using all attention heads}
\label{tab:means_by_percentile}
\end{table}

\subsubsection{Hyperparameter Analysis}
\label{subsub:hyperparameters_analysis}

\name{} introduces two hyperparameters: the gradient flow threshold $\gamma$ and the flow-sparsity balance coefficient $\alpha$. We evaluate the impact of these hyperparameters on the ImageNet1K validation set using the ViT-Base model.

The threshold $\gamma$ (Equation \ref{eq:grad-flow}) filters attention heads based on their relative gradient flow. As shown in Table \ref{tab:Hyperparameter-gamma}, the optimal performance is achieved at $\gamma = 0.25$, with the highest scores in both the Insertion - Deletion metric and the Faithfulness Violation Test across all strategies. Without this filtering ($\gamma = 0$), the Insertion - Deletion score decreases by up to 17.78\%, demonstrating the importance of gradient filtering for noise reduction. Conversely, progressively increasing $\gamma$ results in a steady decline in performance. This confirms that overly aggressive filtering discards secondary attention heads that provide essential semantic information. Thus, $\gamma = 0.25$ strikes the optimal balance between signal fidelity and noise suppression.

\begin{table}[ht!]
\centering
    \resizebox{0.45\linewidth}{!}{%
    \begin{tabular}{c|ccc|ccc}
        \toprule
        & \multicolumn{3}{c}{{\em Insertion - Deletion} ↑} & \multicolumn{3}{c}{{\em Violation Test} ↓} \\
        \cmidrule(l){2-7}
        {\em $\gamma$} & {\em Mean} & {\em Black} & {\em Rand} & {\em Mean} & {\em Black} & {\em Rand} \\
        \midrule
        0 & 0.37 & 0.37 & 0.37 & 0.26 & 0.25 & 0.27 \\
        0.25 & \textbf{0.45} & \textbf{0.44} & \textbf{0.45} & \textbf{0.11} & \textbf{0.12} & \textbf{0.14} \\
        0.50 & 0.38 & 0.37 & 0.37 & 0.23 & 0.24 & 0.22 \\
        0.75 & 0.35 & 0.35 & 0.35 & 0.26 & 0.27 & 0.25 \\
        1 & 0.32 & 0.32 & 0.32 & 0.26 & 0.28 & 0.27 \\
        \bottomrule
    \end{tabular}
    }
    \caption{Values of the Insertion - Deletion metric and the Faithfulness Violation Test returned by \name{} for different values of $\gamma$ using ViT-Base}
    \label{tab:Hyperparameter-gamma}
\end{table}

The flow-sparsity coefficient $\alpha$ (Equation \ref{eq:head-weight}) balances gradient-based flow ($\tilde{g}_q^{(l)}$) and Gini-based sparsity ($\tilde{s}_q^{(l)}$). As reported in Table \ref{tab:Hyperparameter-alpha}, the optimal configuration occurs at $\alpha = 0.50$, where the two metrics are equally weighted. Relying exclusively on sparsity ($\alpha = 0$) results in a significant performance decrease of up to 26.67\% in the Insertion - Deletion metric, suggesting that attention alone lacks a relevant signal for the task. Similarly, using only gradient weighting ($\alpha = 1.0$) results in a 22.22\% decrease. This demonstrates that, without sparsity constraints, diffuse and noisy attention patterns persist. This symmetric degradation at both extremes confirms that gradient flow and Gini sparsity encode complementary information, and their equal combination maximizes explanation faithfulness.

\begin{table}[ht!]
\centering
    \resizebox{0.45\linewidth}{!}{%
    \begin{tabular}{c|ccc|ccc}
        \toprule
        & \multicolumn{3}{c}{{\em Insertion - Deletion} ↑} & \multicolumn{3}{c}{{\em Violation Test} ↓} \\
        \cmidrule(l){2-7}
        {\em $\alpha$} & {\em Mean} & {\em Black} & {\em Rand} & {\em Mean} & {\em Black} & {\em Rand} \\
        \midrule
        0 & 0.33 & 0.34 & 0.33 & 0.25 & 0.26 & 0.24 \\
        0.25 & 0.39 & 0.38 & 0.39 & 0.16 & 0.17 & 0.18 \\
        0.50 & \textbf{0.45} & \textbf{0.44} & \textbf{0.45} & \textbf{0.11} & \textbf{0.12} & \textbf{0.14} \\
        0.75 & 0.42 & 0.41 & 0.42 & 0.15 & 0.16 & 0.17 \\
        1 & 0.35 & 0.36 & 0.36 & 0.24 & 0.26 & 0.28 \\
        \bottomrule
    \end{tabular}
    }
    \caption{Values of the Insertion - Deletion metric and the Faithfulness Violation Test returned by \name{} for different values of $\alpha$ using ViT-Base}
    \label{tab:Hyperparameter-alpha}
\end{table}

\subsection{Qualitative Results}
\label{sub:Qualitative}

In this section, we present the qualitative results from our experimental campaign. Specifically, Subsection \ref{subsub:Classification} focuses on classification, while Subsection \ref{subsub:Segmentation} concentrates on segmentation. 

\subsubsection{Classification}
\label{subsub:Classification}

\begin{figure}[ht!]
    \centering
    \includegraphics[width=1\linewidth]{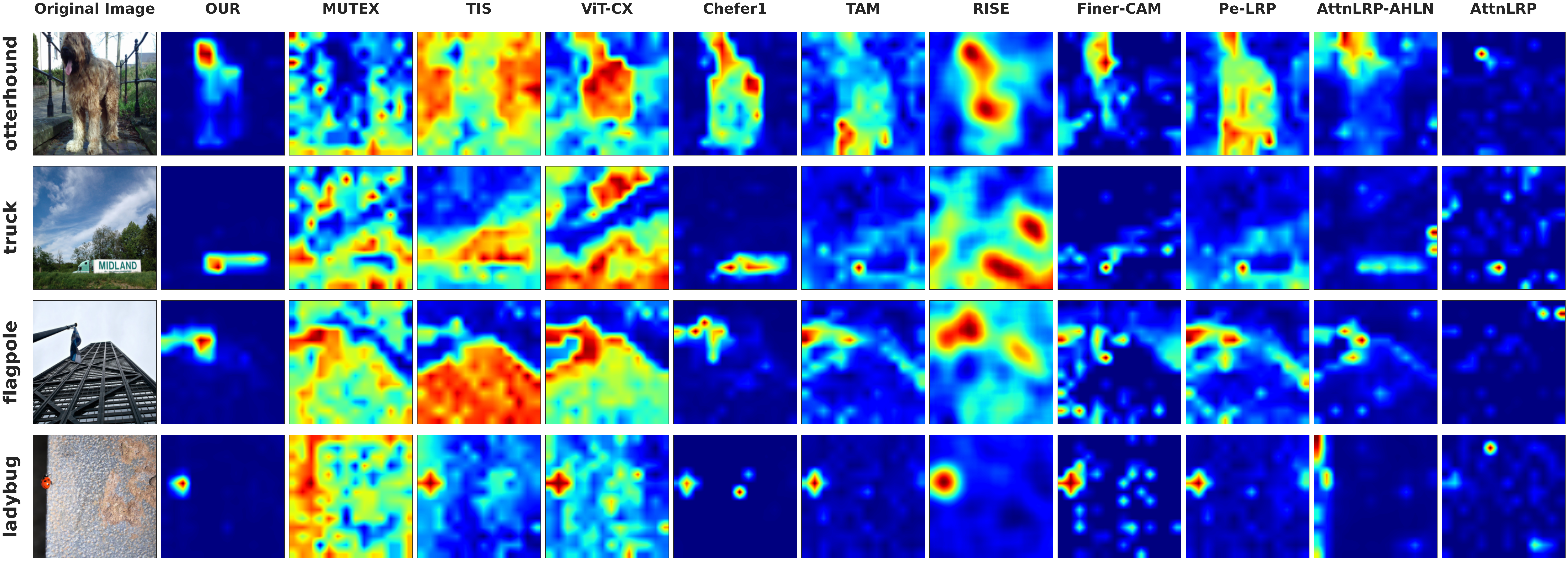}
    \caption{Comparison of the heatmaps returned by \name{} and related approaches for four ImageNet1K images using ViT-Base}
    \label{fig:Comparison-Methods}
\end{figure}

As a first qualitative analysis of \name{} for classification, we compare our framework with related approaches. Figure~\ref{fig:Comparison-Methods} shows the heatmaps generated by GradSkip and competing approaches for four ImageNet1K images (\textit{otterhound}, \textit{truck}, \textit{flagpole}, and \textit{ladybug}) using a ViT-Base backbone. GradSkip produces activation maps that are focused and spatially precise, concentrating on the most discriminative regions of the target object. For the \textit{otterhound} class, for instance, \name{} correctly highlights the dog's body while suppressing the background. In contrast, methods such as TIS, RISE, and AttnLRP exhibit diffuse or noisy activations spread over irrelevant regions. In the \textit{truck} example, GradSkip isolates the vehicle cab with high precision, whereas several baselines (e.g., ViT-CX and TAM) produce scattered activations with limited spatial coherence. For the \textit{flagpole} example, which is challenging due to the thin and elongated structure of the target, \name{} successfully localizes the object, whereas most competitors fail to produce a well-defined response. 
Finally, for the \textit{ladybug} class, which involves a very small object against a cluttered background, GradSkip generates a tightly concentrated heatmap on the insect, confirming its effectiveness even under extreme scale variations.

\begin{figure}[ht!]
    \centering
    \includegraphics[width=0.9\linewidth]{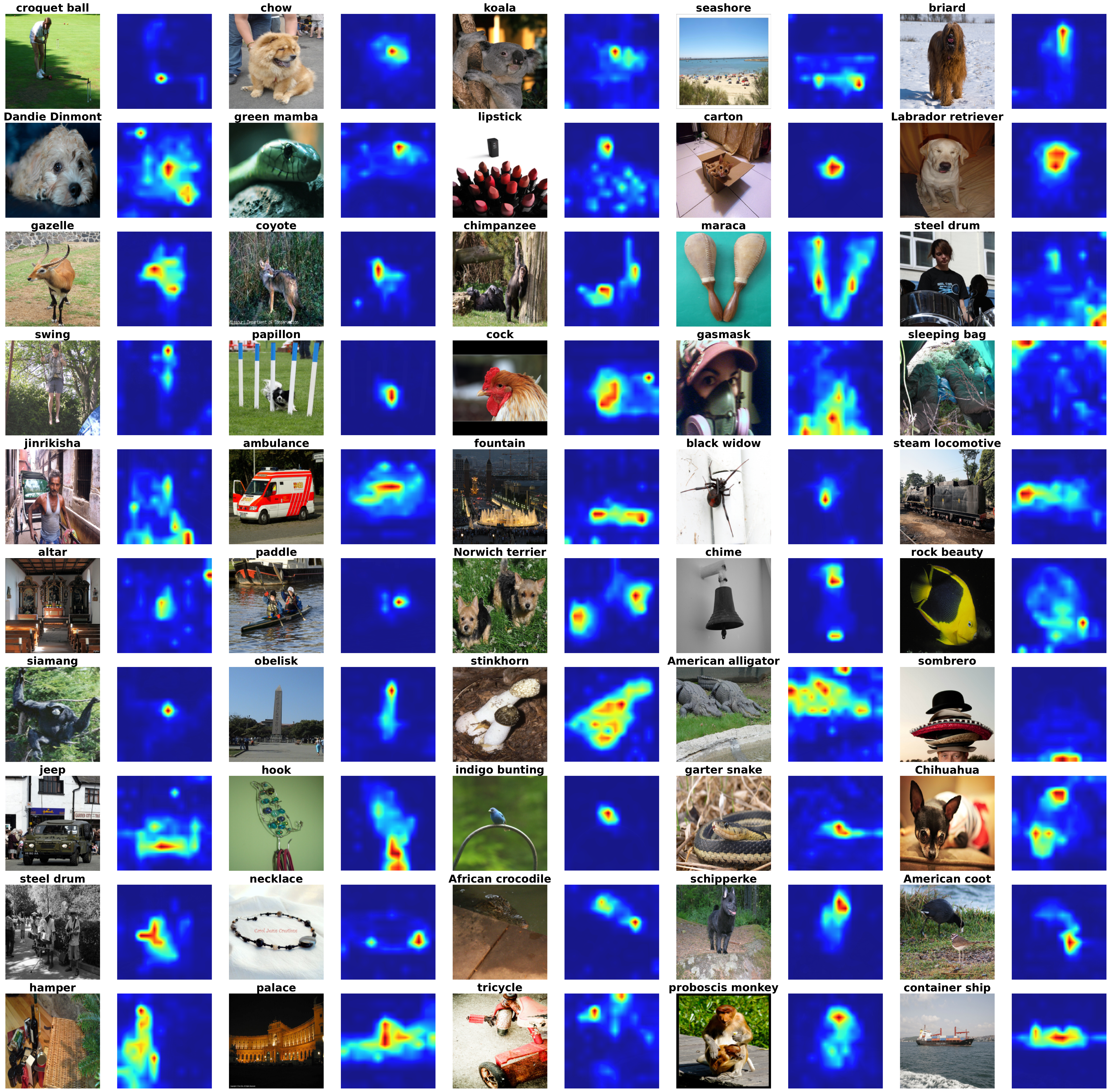}
    \caption{Grid of 50 randomly selected images of ImageNet1K and the corresponding heatmaps produced by \name{} with a ViT-Base backbone}
    \label{fig:compare_classification_grid}
\end{figure}

Figure \ref{fig:compare_classification_grid} displays a grid of 50 randomly selected ImageNet1K images, along with the associated heatmaps returned by \name{}. This figure provides an overview of \name{}'s qualitative behavior across various categories and scene configurations. Its analysis reveals that \name{} produces highly selective heatmaps. This behavior is particularly effective for small objects, as \name{} can often isolate the most discriminative regions with high precision. For larger objects, \name{}'s explanation tends to focus on a subset of highly informative pixels rather than covering the entire object uniformly.

\begin{figure}[ht!]
    \centering
    \begin{subfigure}{\linewidth}
        \centering
        \includegraphics[width=0.75\linewidth]{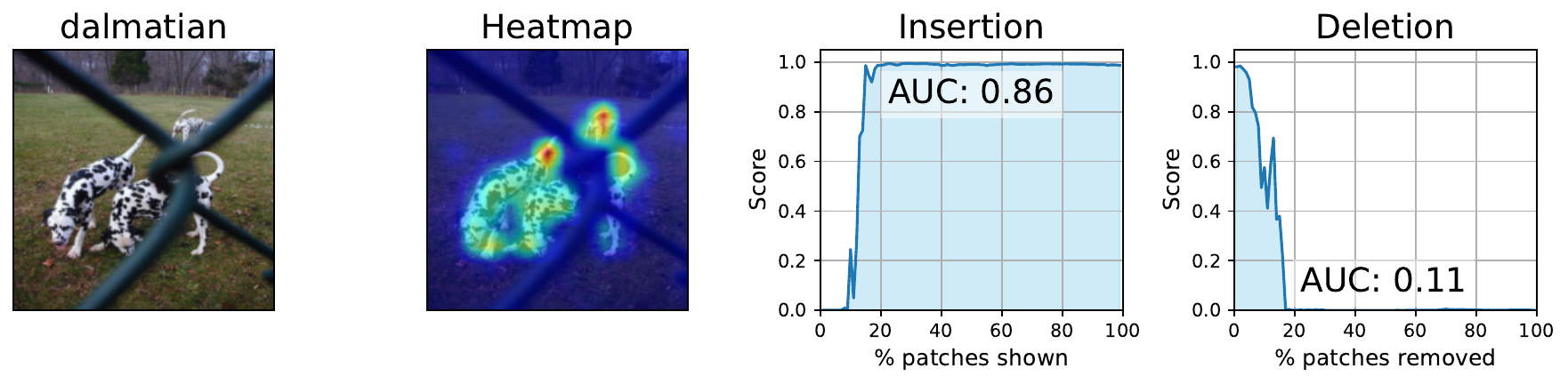}
        \caption{Class \textit{dalmatian}}
        \label{fig:AUC-dalmatian}
    \end{subfigure}

    \vspace{2mm}

    \begin{subfigure}{\linewidth}
        \centering
        \includegraphics[width=0.75\linewidth]{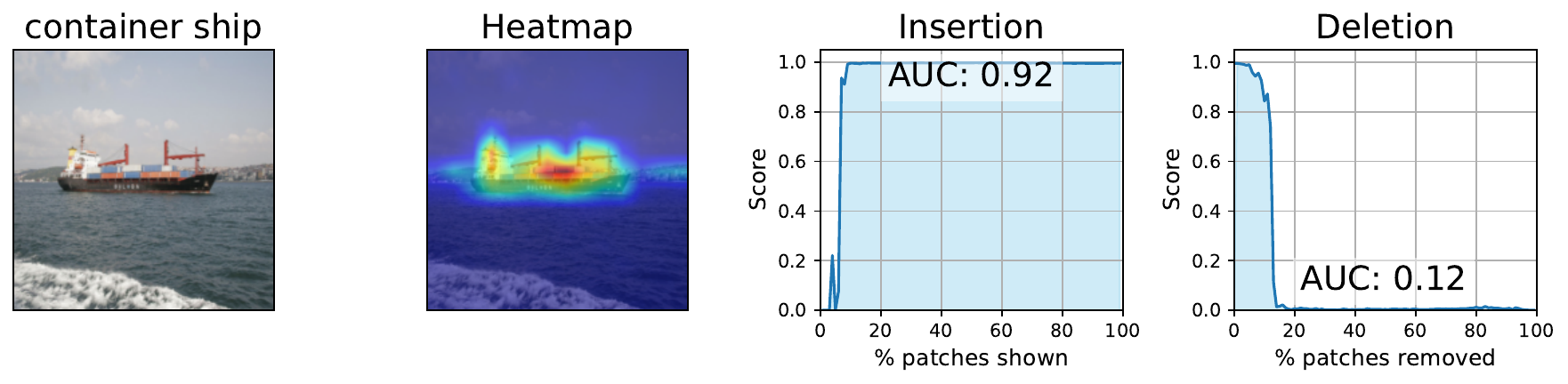}
        \caption{Class \textit{container ship}}
        \label{fig:AUC-containership}
    \end{subfigure}

    \vspace{2mm}

    \begin{subfigure}{\linewidth}
        \centering
        \includegraphics[width=0.75\linewidth]{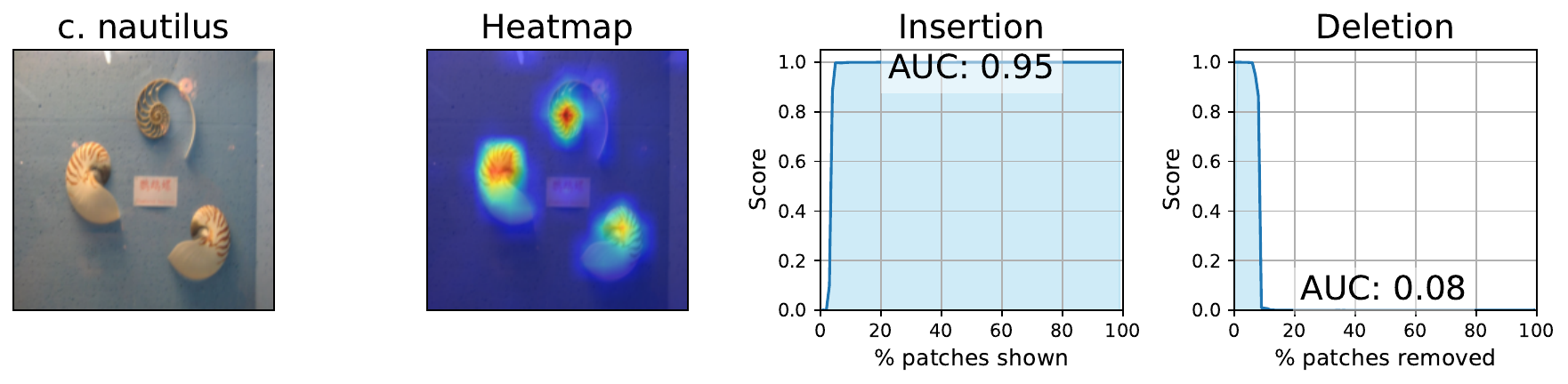}
        \caption{Class \textit{chambered nautilus}}
        \label{fig:AUC-nautilus}
    \end{subfigure}   

    \caption{Heatmap, Insertion curves, and Deletion curves obtained by \name{} for three ImageNet1K images of different classes}
    \label{fig:Qualitative-Comparison-ViT}
\end{figure}

After conducting an initial qualitative analysis of \name{}'s behavior, we proceed with a more in-depth analysis. For this task, we pair each heatmap with the corresponding faithfulness curve. Figure~\ref{fig:Qualitative-Comparison-ViT} presents three examples selected to illustrate \name{}'s behavior alongside its corresponding Insertion and Deletion curves. \Cref{fig:AUC-dalmatian} shows an image of the class {\em dalmatian}, a single-object scene complicated by a distracting foreground. Here, \name{} concentrates its activation on the two dogs while ignoring the metallic fence and grassy background. The steep Insertion curve, together with the rapidly decreasing Deletion curve, confirms that the highlighted regions are sufficient and necessary for prediction. \Cref{fig:AUC-containership} shows an image of the class {\em container ship}, a cleaner single-object scene. Here, the heatmap precisely delineates the vessel and disregards the surrounding water. The similarly steep Insertion and Deletion trends confirm that the highlighted regions are sufficient and necessary for prediction. Finally, \Cref{fig:AUC-nautilus} shows an image of the class {\em chambered nautilus}, a more challenging case in which three instances of the same class coexist in the scene. In this setting, \name{} correctly localizes all three instances against the uniform background, confirming its ability to capture class-relevant evidence regardless of object multiplicity. Across these increasingly difficult settings, \name{} consistently produces heatmaps with three desirable properties: high spatial precision in localizing discriminative regions, minimal background noise, and semantic coherence aligned with human expectations.

To further explore how \name{} behaves with respect to Insertion and Deletion metrics, we calculate the Insertion and Deletion AUC curve values for three groups of ImageNet1K images: namely easy classes, hard classes, and multi-object scenes. These groups are defined by the visual complexity of the images. Easy examples typically contain a dominant foreground object and limited background clutter. Hard examples involve more complex context or stronger visual interference. Multi-object examples include multiple relevant semantic instances.

\begin{figure}[ht!]
    \centering
    \includegraphics[width=0.7\linewidth]
    {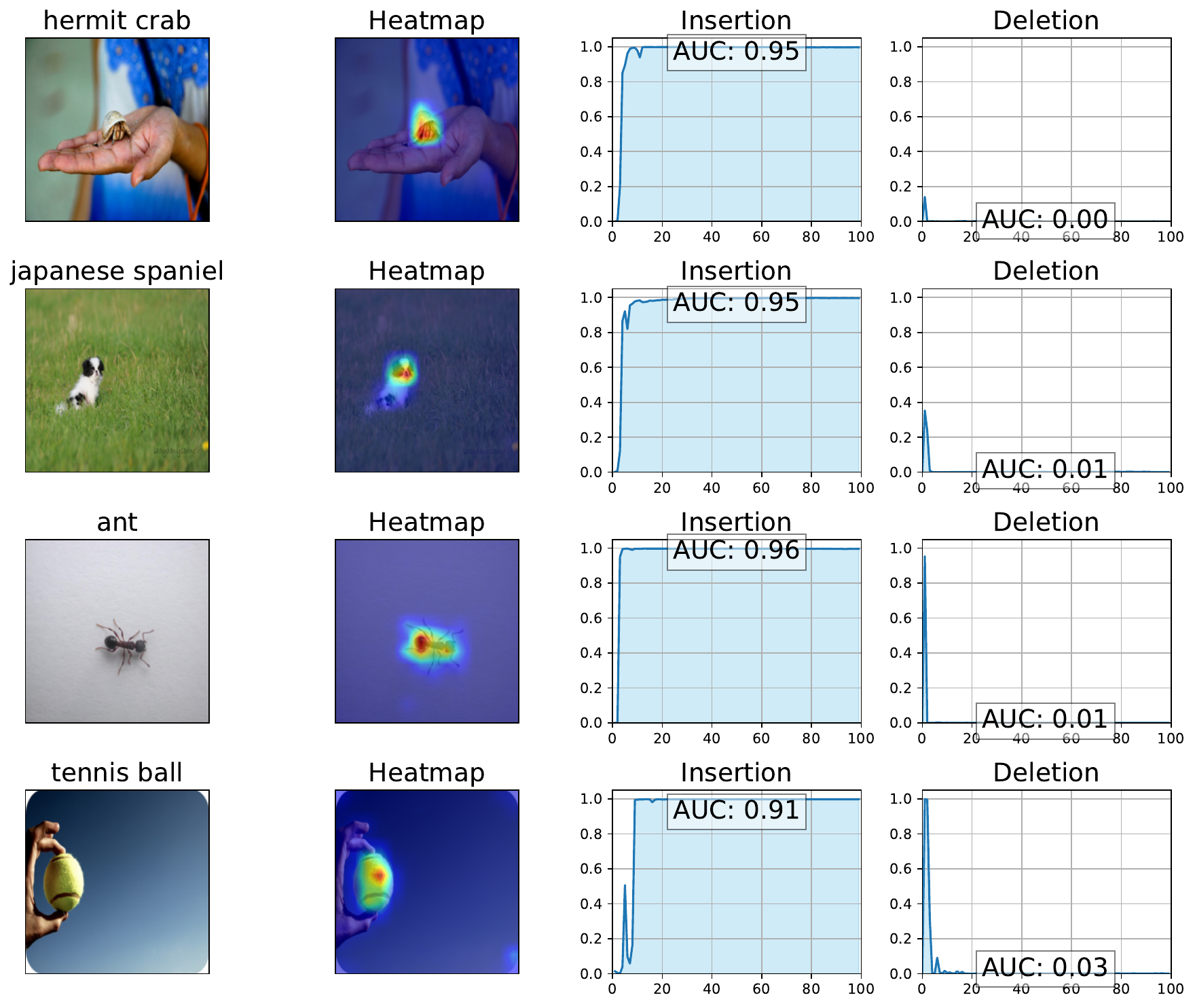}
    \caption{Insertion and Deletion AUC curves for easy ImageNet1K examples using a ViT-Base backbone}
    \label{fig:compare_classification_easy}
\end{figure}

For the easy examples (\Cref{fig:compare_classification_easy}), the curves confirm \name{}'s strong performance, with very high Insertion AUC values and very low Deletion AUC values across the reported examples. This suggests that \name{} identifies compact and highly discriminative regions that strongly support target predictions.

\begin{figure}[ht!]
    \centering
    \includegraphics[width=0.7\linewidth]
    {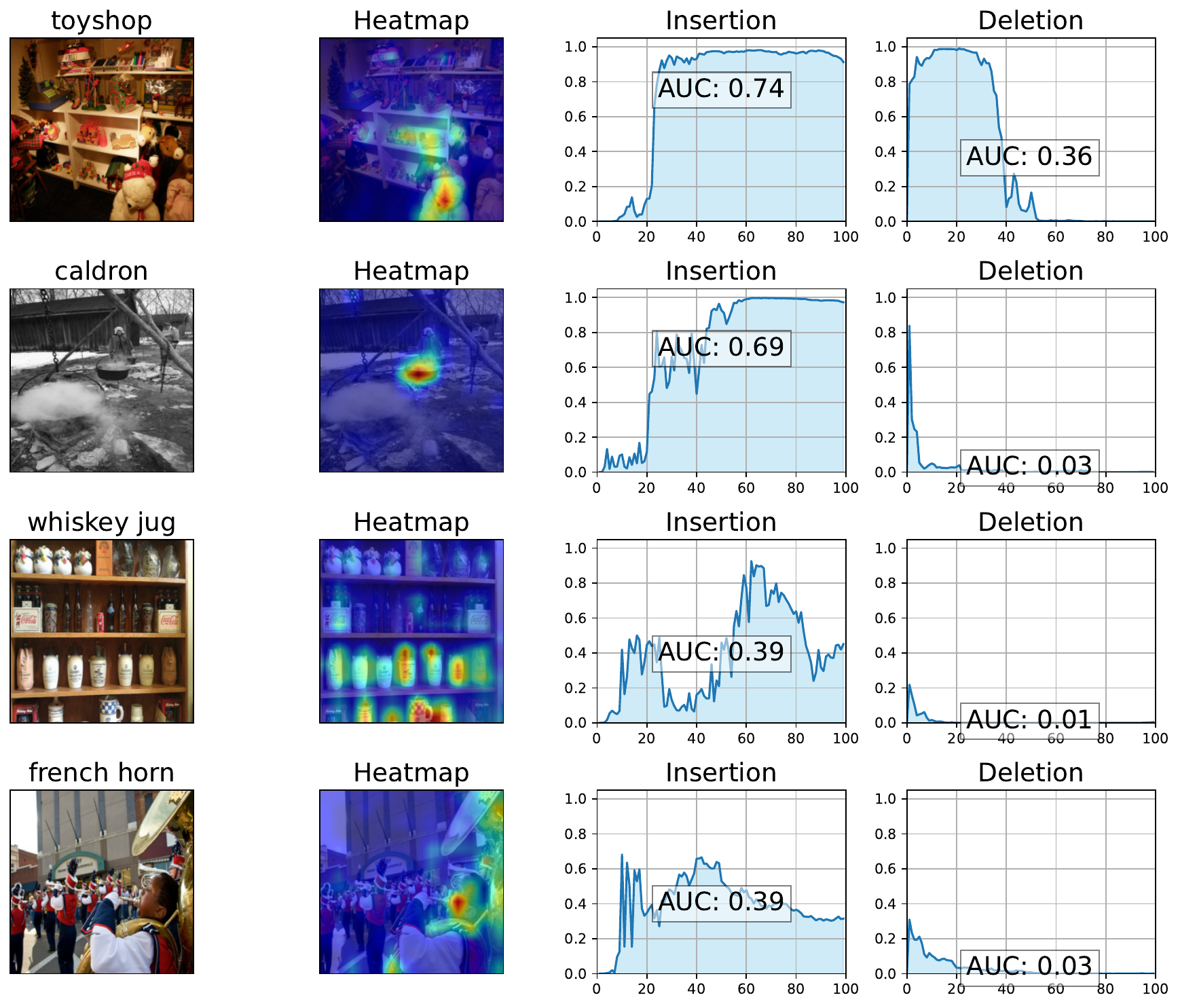}
    \caption{Insertion and Deletion AUC curves for hard ImageNet1K examples using a ViT-Base backbone}
    \label{fig:compare_classification_hard}
\end{figure}

For the hard examples (\Cref{fig:compare_classification_hard}), \name{} still performs well, although the absolute AUC values are clearly lower because the images are more ambiguous and visually complex. A representative case is the image {\em whiskey jug}, where \name{} highlights the relevant object regions, but the ViT prediction on the full image has a confidence level of about 50\%, resulting in relatively low Insertion--Deletion values despite a meaningful qualitative explanation.

\begin{figure}[ht!]
    \centering
    \includegraphics[width=0.7\linewidth]{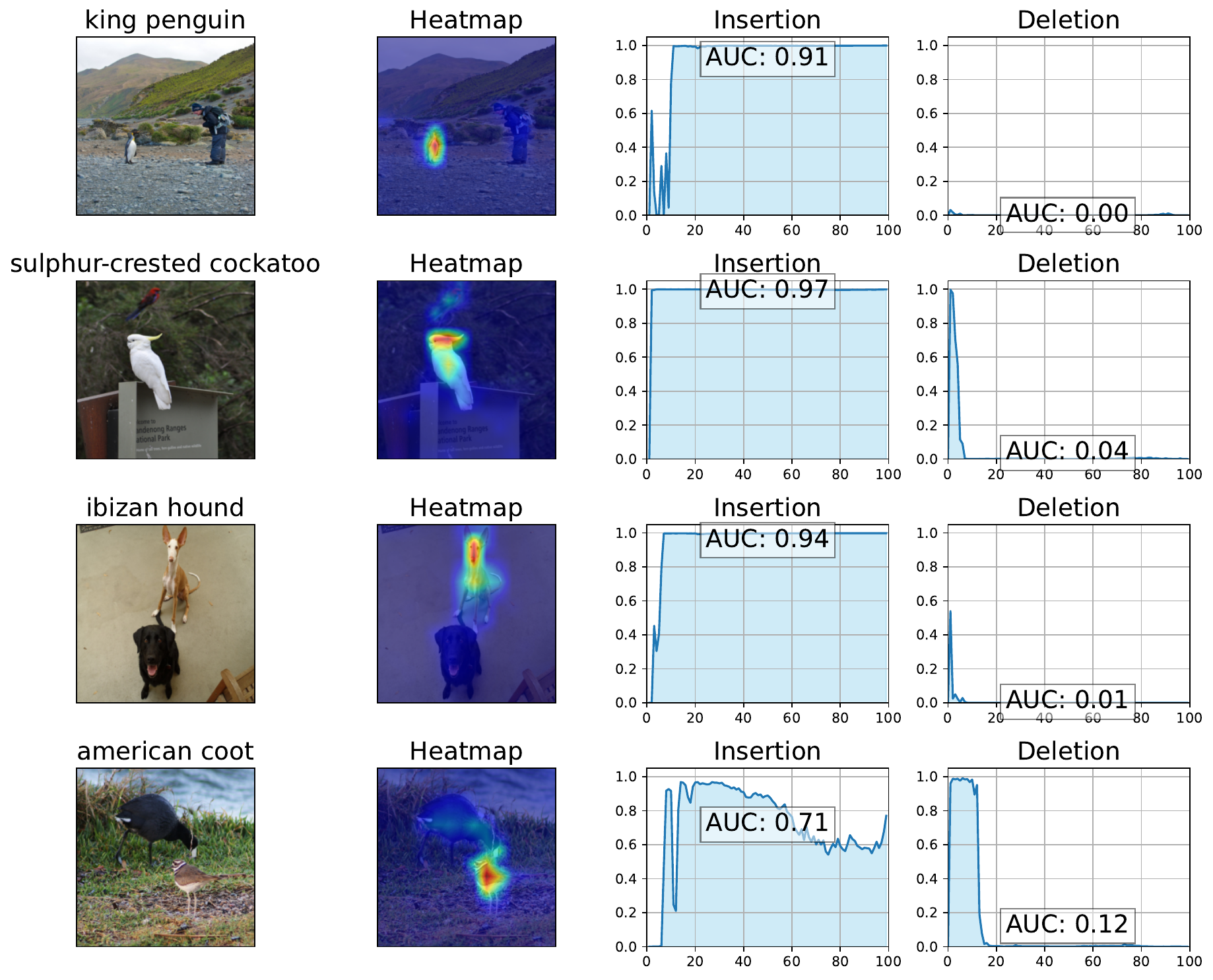}
    \caption{Insertion and Deletion AUC curves for multi-object ImageNet1K scenes using a ViT-Base backbone}
    \label{fig:compare_classification_multi}
\end{figure}

The multi-object examples (\Cref{fig:compare_classification_multi}) further demonstrate that \name{} remains effective even when several semantic elements coexist in the same scene. In these cases, \name{} achieves high Insertion AUC values and maintains low Deletion AUC values on most examples. This suggests that the selected regions are closely tied to the model's decision.

\subsubsection{Segmentation}
\label{subsub:Segmentation}

In this section, we present a qualitative analysis of the heatmaps generated by \name{} for the segmentation task. 

\begin{figure}[ht!]
    \centering
        \includegraphics[width=0.6\linewidth]{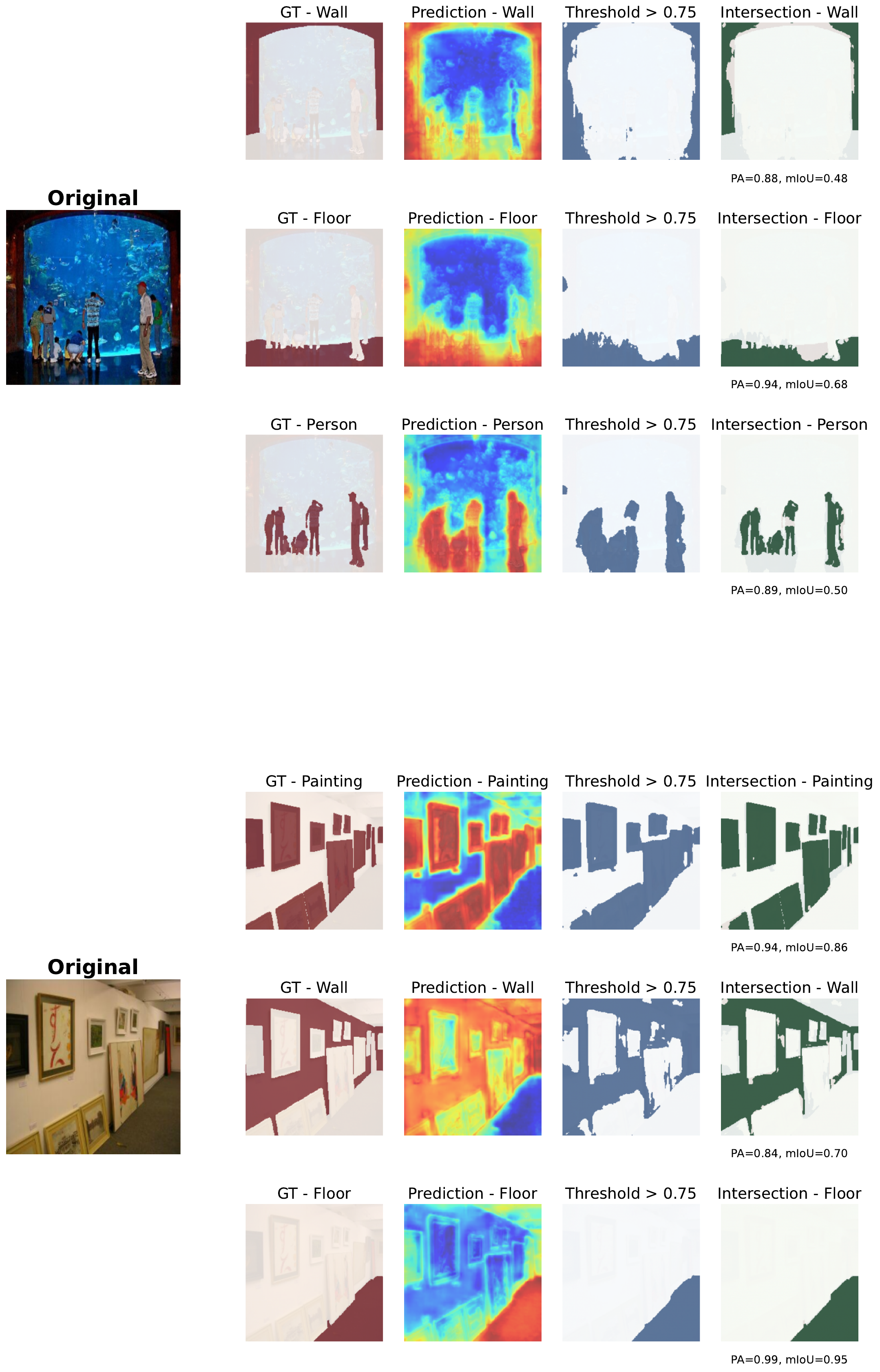}
        \caption{Qualitative segmentation results of \name{} on two ADE20K-150 images using SegFormer. For each target class, we present the original image, the ground truth mask, the predicted relevance map, the thresholded map, and the intersection of the thresholded map and the ground truth}
        \label{fig:Qualitative-Comparison-SegFormer}
\end{figure}

Figure~\ref{fig:Qualitative-Comparison-SegFormer} illustrates our qualitative analysis of the semantic segmentation task using the ADE20K-150 dataset. For these analyses, we employed the SegFormer model and two images containing different classes. The figure displays the heatmaps generated by \name{}, the corresponding ground truth, and their intersection. \name{} demonstrates a remarkable ability to dynamically provide the correct heatmap for different target classes. The generated heatmaps closely align with the underlying ground truth shapes, effectively isolating the desired class instances while disregarding the rest of the scene. These visual results strongly corroborate the quantitative findings in Table~\ref{tab:segmentation_results}, confirming that \name{} prevents over-assignment to background regions and precisely delineates class-specific areas, even in complex, multi-object environments.

\begin{figure}[ht!]
    \centering
    \includegraphics[width=0.6\linewidth]{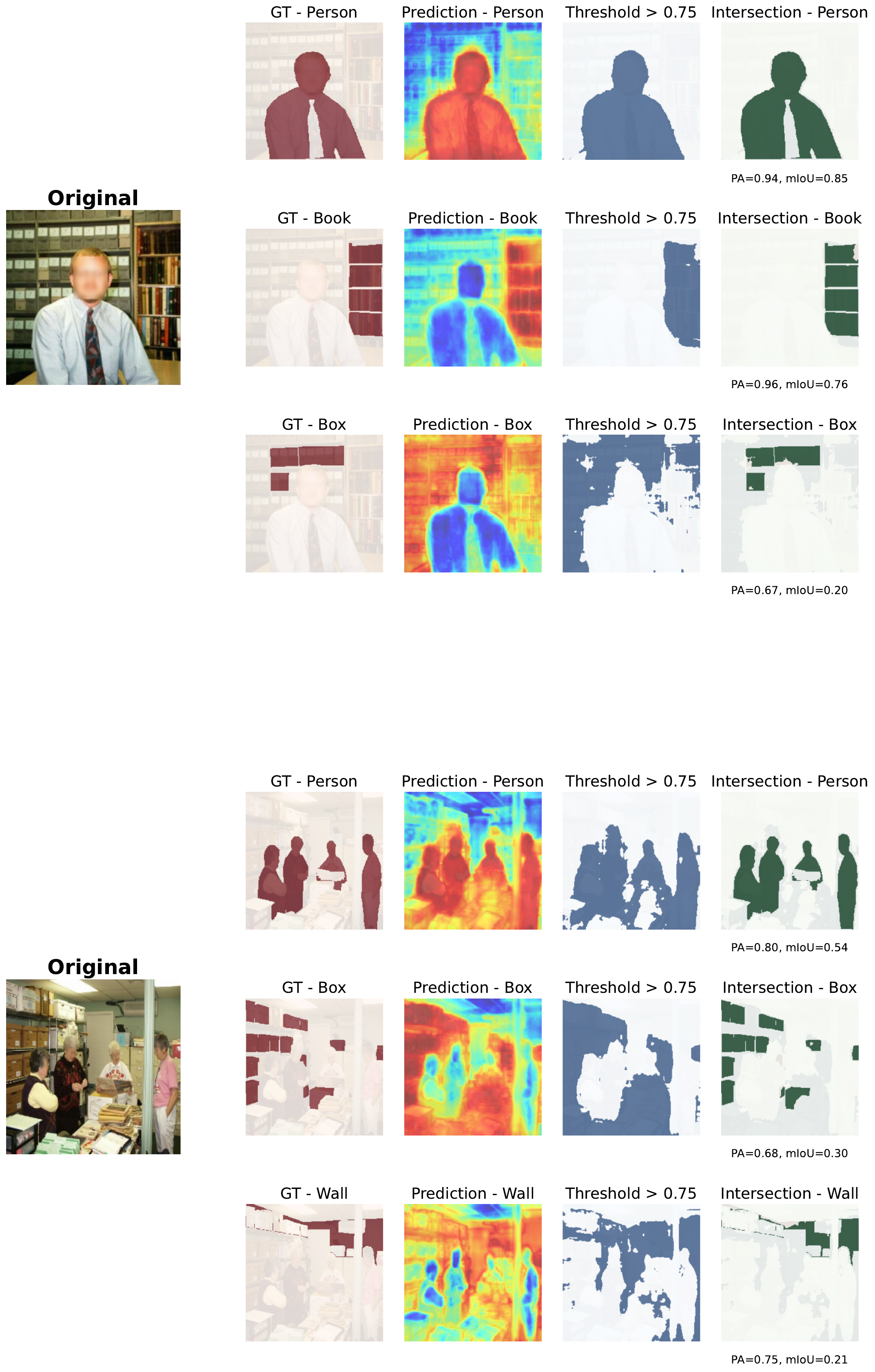}
        \caption{A further qualitative segmentation test on ADE20K-150 using SegFormer}
        \label{fig:compare_segmentation}
\end{figure}

\begin{figure}[ht!]
    \centering
    \includegraphics[width=0.6\linewidth]{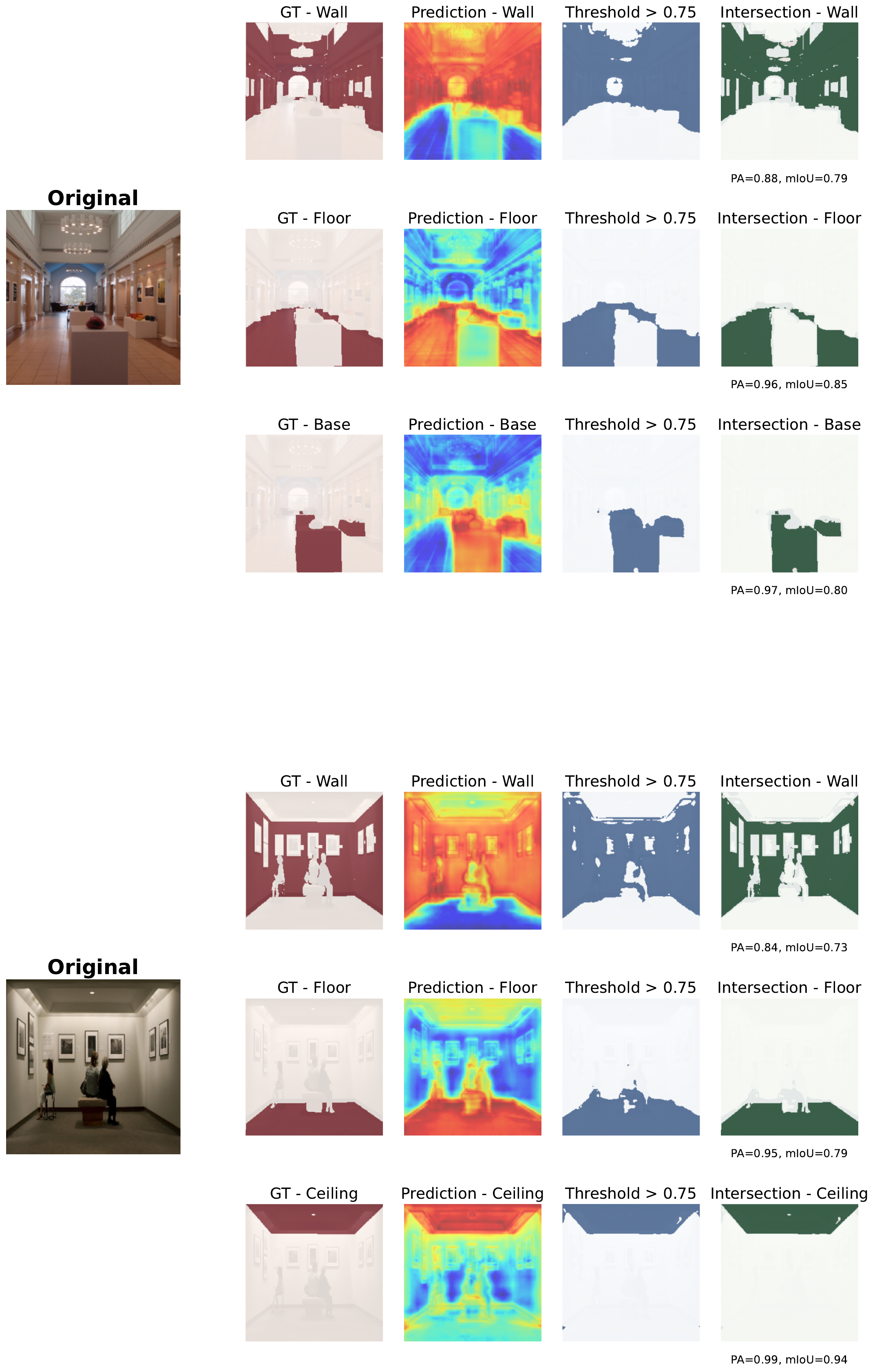}
    \caption{A final qualitative segmentation test on ADE20K-150 using SegFormer}
        \label{fig:compare_segmentation-1}
\end{figure}

We continue our analysis with two additional tests, where we apply \name{} with SegFormer to some images from the ADE20K-150 dataset. \Cref{fig:compare_segmentation,fig:compare_segmentation-1} show the results obtained. For each target class, we present the original image, the ground truth mask, the predicted relevance map, the thresholded map (i.e., pixels with saliency greater than $0.75$), and the intersection of the thresholded map and the ground truth. These visualizations demonstrate \name{}'s consistent, class-specific localization capability across various scenes and categories. More specifically, a  stronger alignment can be observed for broad, spatially coherent classes, such as wall, floor, ceiling, person, and book. Smaller or less regular categories, such as table and box, remain more challenging to align.

\section{Discussion}
\label{sec:Discussion-section}

In this section, we examine the reasons behind \name{}'s behavior and the most important design choices underlying it. \name{} is motivated by two architectural assumptions: the heterogeneous importance of attention heads and the explicit role of skip connections. These assumptions are reflected in the empirical findings. On ImageNet1K, \name{} outperforms MUTEX (albeit MUTEX was already strong) by up to 12.50\% in the Insertion - Deletion metric and improves stability in the Faithfulness Violation Test by 21.43\%. These gains carry over to the medical domain of BloodMNIST. The ablation study clarifies that neither signal alone is sufficient to explain these improvements. Relying solely on sparsity or gradient flow degrades the Insertion - Deletion score by up to 26.67\% and 22.22\%, respectively. Disabling gradient filtering reduces the score by up to 17.78\%. These results support our central claim that faithful attribution to transformers must account for the functional specialization of heads, rather than aggregating them uniformly.

The selectivity that drives this faithfulness becomes even more apparent when the explanations are evaluated spatially. On the Pointing Game metric, \name{} essentially performs as well as the strongest baselines. However, on segmentation, \name{} exceeds MUTEX by 53.94\% in mIoU on ADE20K-150, creating a significant margin. This difference stems from the nature of the two tasks. The Pointing Game metric only rewards hitting the most relevant point. Several methods can reach this target, even when their maps are diffuse, so the metric is relatively forgiving. Conversely, segmentation penalizes over-extended maps through the union term of the mIoU. Concentrating relevance on truly discriminative regions pays off precisely here. A comparison with Finer-CAM is illustrative. Finer-CAM achieves comparable Pixel Accuracy but a much lower mIoU than \name{}. In fact, \name{} outperforms Finer-CAM by 256.73\% because Finer-CAM's maps spread saliency over large portions of the image. Thus, \name{}'s high faithfulness stems from its ability to sharply isolate relevant regions rather than from broad coverage.

This quality can be achieved without an expensive procedure. \name{} relies on a single forward and backward pass. It requires only 69.39 GFLOPs for classification and 28.52 GFLOPs for segmentation. This makes \name{} comparable to LRP-based approaches and far more efficient than perturbation-based approaches, such as MUTEX and TIS. Unlike the cheapest baselines, however, this efficiency does not come at the expense of explanation quality. LRP variants that match or slightly exceed the GFLOPs required by \name{} consistently produce weaker explanations. \name{} outperforms the best of these methods by 41.37\% in the Pointing Game metric on ViT-Base. Therefore, it provides a favorable trade-off between explanation quality and computational cost. This makes \name{} suitable for settings where attribution must be computed repeatedly or with limited resources.

These observations also define the limits of \name{}. The previously discussed advantages, such as high faithfulness across architectures and datasets, sharp delineation of class-specific regions, and low computational cost, depend on the gradients and activations of the explained model. Since the attribution process requires a backward pass, \name{} cannot be applied in purely forward, gradient-free settings. \name{} also requires access to the model internals. Additionally, \name{} only propagates relevance through patch tokens, discarding the CLS token. This makes it directly applicable to ViT variants without a CLS token, such as SegFormer. However, in standard ViTs, the CLS token is used to compute the prediction and aggregate global context from patches at every layer. Therefore, ignoring it during backpropagation may discard useful class-relevant signals on architectures where the CLS plays a central role.

\section{Conclusion}
\label{sec:Conclusion}

This paper introduced \name{}, an explainability method for ViTs that overcomes the limitations of existing relevance-propagation approaches. Specifically, \name{} considers the varying importance of attention heads using adaptive weighting and incorporates skip-aware relevance propagation to model the contribution of residual connections.

In a broad range of experiments, \name{} achieved state-of-the-art performance in terms of faithfulness for the classification task on both the ImageNet1K and BloodMNIST datasets. It was also equally effective for semantic segmentation on ADE20K-150, achieving the highest Pixel Accuracy and mIoU among competing approaches. These results were obtained at a computational cost comparable to the most efficient propagation approaches and orders of magnitude lower than that of perturbation-based approaches, offering a favorable trade-off between explanation quality and efficiency.

There are several promising avenues for future work. First, adapting the head weighting strategy to sparse or linearized attention variants, such as those adopted in efficient transformers for high-resolution inputs, is a natural architectural extension. Second, rather than relying on the fixed hyperparameter $\alpha$ (balancing causal relevance and spatial coherence - see Equation \ref{eq:head-weight}), exploring learnable fusion strategies for flow sparsity combinations could improve the adaptability of the head weighting mechanism across different architectures and tasks. Third, extending \name{} to video transformers could broaden its applicability to video understanding tasks, such as action recognition. Achieving this would require incorporating temporal dependencies across frames in the relevance propagation process. 

\section*{Declaration of competing interest}
The authors declare that they have no known competing financial interests or personal relationships that could have appeared to influence the work reported in this paper.

\section*{Data availability}
The code used for our study will be released after publication.

\bibliographystyle{plain}
\bibliography{bibliografia}

\end{document}